\documentclass[letterpaper, 10 pt, conference]{ieeeconf}  

\usepackage[utf8]{inputenc}
\usepackage{graphicx}
\usepackage{booktabs}
\usepackage{cite}
\usepackage{xcolor}
\graphicspath{{image/}}
\usepackage{etoolbox}
\makeatletter
\patchcmd{\@makecaption}
  {\scshape}
  {}
  {}
  {}
\makeatletter
\patchcmd{\@makecaption}
  {\\}
  {.\ }
  {}
  {}
\makeatother

\IEEEoverridecommandlockouts                              

\usepackage{makecell}
\newcommand{\tabincell}[2]{\begin{tabular}{@{}#1@{}}#2\end{tabular}} 
\newlength\savewidth

\definecolor{citecolor}{HTML}{0071BC}
\usepackage[colorlinks]{hyperref}
\hypersetup{colorlinks,linkcolor={red},citecolor={citecolor}}  

\begin{document}

\title{\LARGE \bf
RSRD: A Road Surface Reconstruction Dataset and Benchmark \\ for Safe and Comfortable Autonomous Driving
}

\author{Tong Zhao$^{1}$, Chenfeng Xu$^{2}$, Mingyu Ding$^{2*}$, Masayoshi Tomizuka$^{2}$, Wei Zhan$^{2}$, and Yintao Wei$^{1*}$
\thanks{$^{1}$School of Vehicle and Mobility, Tsinghua University, Beijing, China
        {\tt\small zhaot20@mails.tsinghua.edu.cn; weiyt@tsinghua.edu.cn}}%
\thanks{$^{2}$ Department of Mechanical Engineering, University of California Berkeley, CA, USA
        {\tt\small \{xuchenfeng, myding, tomizuka, wzhan\}@berkeley.edu}}%
\thanks{The work was partially done during Tong Zhao's visit to UC Berkeley.}
\thanks{* Corresponding authors.}
}

\maketitle
\thispagestyle{empty}
\pagestyle{empty}

\begin{abstract}



%
This paper addresses the growing demands for safety and comfort in intelligent robot systems, particularly autonomous vehicles, where road conditions play a pivotal role in overall driving performance. 
For example, reconstructing road surfaces helps to enhance the analysis and prediction of vehicle responses for motion planning and control systems.
We introduce the Road Surface Reconstruction Dataset (RSRD), a real-world, high-resolution, and high-precision dataset collected with a specialized platform in diverse driving conditions.
It covers common road types containing approximately 16,000 pairs of stereo images, original point clouds, and ground-truth depth/disparity maps, with accurate post-processing pipelines to ensure its quality.
Based on RSRD, we further build a comprehensive benchmark for recovering road profiles through depth estimation and stereo matching.
Preliminary evaluations with various state-of-the-art methods reveal the effectiveness of our dataset and the challenge of the task, underscoring substantial opportunities of RSRD as a valuable resource for advancing techniques, \emph{e.g.}, multi-view stereo towards safe autonomous driving.
The dataset and demo videos are available at \href{https://thu-rsxd.com/rsrd/}{https://thu-rsxd.com/rsrd/}.

%
%
%
%


\end{abstract}

\section{INTRODUCTION}

As one of the most potential application scenarios of computer vision and artificial intelligence, environmental perception has laid the foundation for subsequent motion planning and control systems of unmanned robots and ground vehicles~\cite{10089400}~\cite{8715479}.
The realm of autonomous vehicles (AVs) has been under study for decades, and in recent years, remarkable strides 
on both algorithms and datasets~\cite{9380166}~\cite{9173706} have been introduced for primary motion control on above-road traffic scenarios.
However, existing AV research typically focuses on macro-level traffic situations. For example, semantic segmentation, detection, and tracking pipelines are designed for transportation-level applications such as collision avoidance, lane change, and cruise.
The micro-level traffic situations, e.g., the road surface conditions, are rarely considered.
%

The road surface conditions, particularly the friction and unevenness parameters, are frequently overlooked or simplistically treated as constant constraints within planning and control systems.
In this case, the precision and performance of control systems are inherently limited, given that the actual dynamic characteristics of the controlled objects are unknown.
The road surface, being the sole interface with which vehicles establish physical contact, essentially defines the safety and comfort boundaries of vehicle dynamics~\cite{10.1115/1.4047962}~\cite{THEUNISSEN2021206}, \emph{i.e.}, every force affecting the vehicle's motion arises from interactions between the tires and the road surface. 
Besides understanding the broader traffic environment, road surface perception/reconstruction remains a critical bottleneck in ensuring overall autonomous driving performance.


RSCD~\cite{10101715} explores the micro-level road conditions by treating it as a classification task.
To capture more detailed surface structures, \cite{7797253}~\cite{9385918} propose to reconstruct the road surface profile.
A finely detailed road profile is crucial for estimating both the tire-road friction coefficient and the road's unevenness. %
Utilizing such road structure and texture information gathered by cameras aids in predicting the vehicle's response in advance, enabling proactive decisions to avoid potential safety risks~\cite{9782570,9830130,DU2022103489}. 
Nevertheless, there is still a significant absence of a real-world, large-scale vision dataset or benchmark specifically designed for road reconstruction purposes.
%
%
Existing datasets are generally captured in cities with structured roads, and the scenario coverage is insufficient for practical road texture perception applications. 
%

\begin{figure*}[t]
    \centering
    \includegraphics[width=1\linewidth]{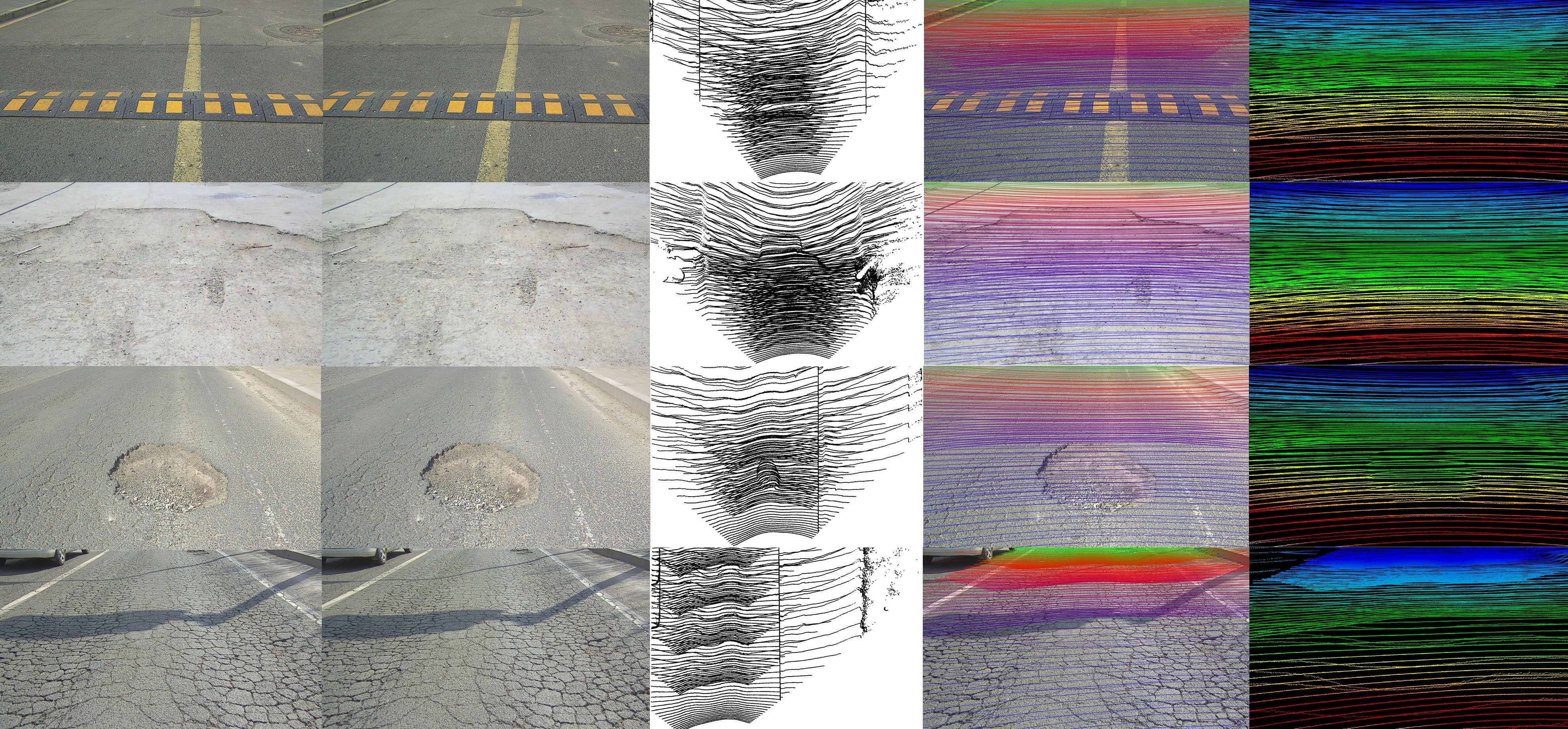}
    \caption{Examples of the RSRD. From left to right: left images, right images, fused point clouds, images with projected point cloud, and disparity maps.}
    \label{example}
\end{figure*}

To solve the above problems, this work presents a comprehensive exploration of fine-grained road surface reconstruction based on visual inputs.
We first create a dataset, called RSRD (see Figure~\ref{example}), which to the best of our knowledge, is the first large-scale and real-world dataset for road surface reconstruction.
It contains calibrated, high-precision images, point clouds, and detailed motion information.
Building upon RSRD, we further establish a benchmark by evaluating various state-of-the-art algorithms for monocular depth estimation and stereo matching tasks.
Additionally, our dataset can serve as an effective resource for tasks encompassing reconstruction, localization, and direct point cloud processing.

We believe our dataset and benchmark represent a pioneering contribution toward autonomous driving safety and comfort, \emph{e.g.}, the longitudinal elevation of the road can be anticipated through the texture reconstruction profiles, benefiting active suspension control systems designed to mitigate vertical vibrations.
Moreover, when seamlessly integrated with our prior work, RSCD~\cite{10101715}, which provides road friction and material labels, it allows for more comprehensive monitoring of road surface conditions and driving environment understanding.



\section{RELATED WORKS}

The progress of AV perception is always promoted by the emergence of large-scale and real-scenario datasets. Extensive image datasets with point cloud labels acquired under various conditions have been published in the past decade. The 3D surroundings can be recovered by deep learning models with image and point cloud data. Road surface perception, especially reconstruction with cameras, is an emerging topic in the technical stacks of AVs \cite{mei2023rome} \cite{ma2022computer}. There are hardly unified and comprehensive datasets to develop and evaluate road reconstruction algorithms. Although many works with existing datasets have been reported, the accuracy and fineness are not sufficient for real-vehicle applications \cite{8300645} \cite{wang20233d}.

\noindent
\textbf{Resolution.} The cameras are generally installed on the vehicle top to capture the whole surroundings, leaving a small area for the road surface. The image definition in the road area is relatively low, especially at far distances because of the perspective effect. The road surface is inherently texture-less, which is further exacerbated by this installation method. Recovering dense and accurate road profiles is quite challenging since the texture details are lost. The KITTI dataset is a milestone in AV perception and is the most commonly utilized benchmark for various evaluation tasks. However, the image resolution (0.5M) and the road area ratio are small. The datasets released in the following years such as Argoverse~\cite{Chang_2019_CVPR}, A2D2~\cite{geyer2020a2d2}, and FordAV Datasets~\cite{agarwal2020ford}, have higher image resolution even to 8M, while still have small road areas. 

\noindent
\textbf{Ground-truth accuracy and density.} Vision-based perception generally relies on ground-truth depth acquired by LiDAR. The accuracy and density of point cloud labels significantly determine the model performance. Unlike traffic objects such as pedestrians and vehicles with large scale, road unevenness such as rocks and cracks generally have small amplitudes~\cite{5371962}. Most of the LiDAR sensors utilized in datasets have an accuracy of ±3cm, which is insufficient to capture accurate road profile variations. The ApolloScape dataset utilizes a single-line LiDAR with ±5mm accuracy but is counteracted by the poor CAD fitting~\cite{wang2019apolloscape}. Motion compensation to point clouds is also crucial to reach high accuracy. Single LiDAR acquires sparse point clouds at the road area, especially at far distances. Multiple LiDARs are gradually equipped to capture as many details as possible, such as PandaSet (2×LiDAR), nuScenes (5×)~\cite{nuscenes}, and Waymo (5×) \cite{Sun_2020_CVPR}. Multi-frame fusion is also conducted to further improve the point cloud density, though it may affect little in the road surface area.

Moreover, the scenario diversity in most datasets is inadequate for real-vehicle reconstruction since they are generally collected in city scenarios and the roads are in good condition. The lack of high dynamic range of cameras (in the early datasets) results in poor imaging quality. Also, motion blur may occur in the road area because of the high relative velocity. To eliminate the limitations above, we strive to improve the quality and accuracy of the RSRD. The hardware platform designated for road perception retains detailed road texture. Dense point cloud labels are established by accurately fusing nearby frames.

\section{RSRD}

\subsection{Hardware Platform}

\begin{figure}[t]
    \centering
    \includegraphics[width=1\linewidth]{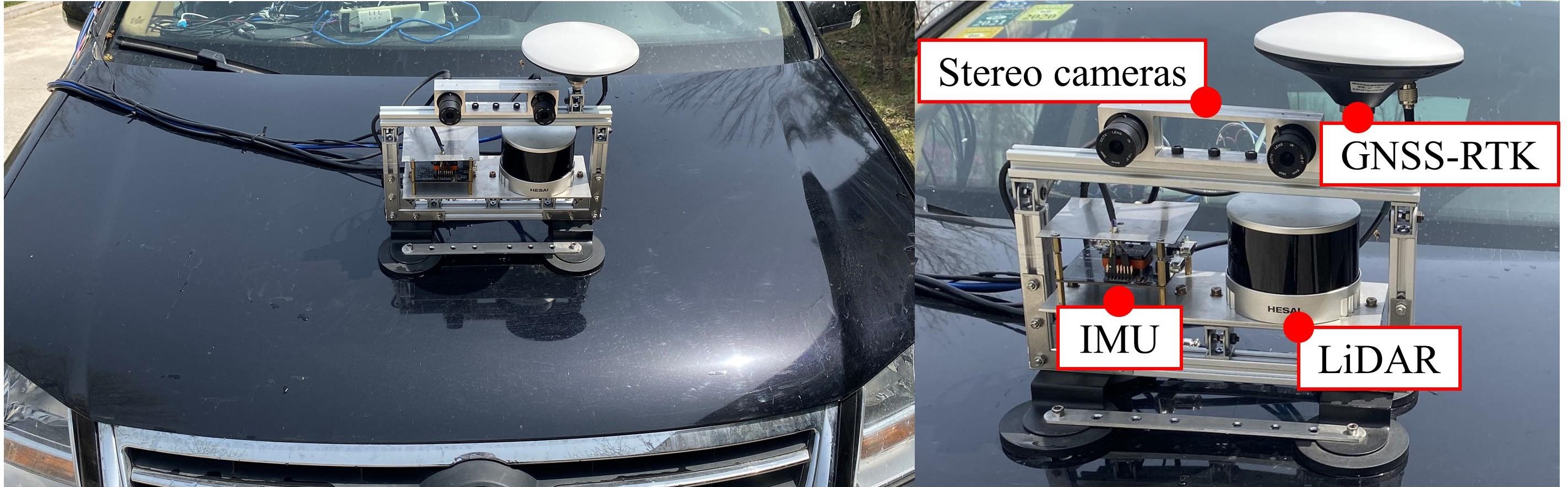}
    \caption{An illustration of our hardware platform.}
    \label{hardware}
\end{figure}

Figure~\ref{hardware} shows the sensor configurations. Unlike the common sensor installation, the suit is mounted on the bonnet and has a certain pitch angle for prototype purposes. The perspective of sensors focuses more on the road area rather than the whole surrounding. The suit consists of an XT32 LiDAR (32-line, 0.09° horizontal resolution), two cameras (LI-AR023ZWDR,1920*1080 resolution, 6mm lens), an IMU (XSENS MTi670, ±0.2° roll and pitch accuracy), and an RTK system (UBlox F9P, 1.4cm horizontal precision, 1cm vertical precision, and ±0.1° heading accuracy). The typical accuracy and precision of the LiDAR are ±1cm and 0.5cm respectively, which are higher than most of these in the existing datasets. It can capture mild road undulations and damages, ensuring high-precision road perception. Since we consider only the road surface area, the horizontal viewing angle of the mechanical rotating LiDAR is set to 100°.

The cameras generate clear and sharp images with up to 105dB, guaranteeing imaging quality in severe brightness changes. Also, it has motion compensation to prevent ghost blur in multi-exposure HDR. The two cameras are fixed by a designed rigid holder with a 12cm baseline. The preview distance of the cameras is about 12m, resulting in disparity values between 20\textasciitilde140. The IMU and the RTK antenna are placed near the LiDAR to measure its orientation and position. We established a temporary fixed basement to achieve more stable and reliable localization results. The position and pose measurements are utilized in the following multi-frame point cloud fusion. The stereo cameras and the camera-LiDAR extrinsic are calibrated with high-precision checkerboards. The re-projection error is smaller than 1 pixel. 

The acquisition system runs at 5Hz, so the LiDAR can acquire more points in one frame. All the sensors are hardware-synchronized by the PPS from GNSS. The cameras start exposure when the LiDAR rotates to the forward position. All the data samples have timestamps. The sensors are integrated with the aluminum profile framework and tightly fixed to ensure a rigid connection. 

\subsection{Data Acquisition and Pre-processing}
The experiments are conducted from March to April, 2023 in Qingdao, China. Driving on uneven roads results in severe vibration of the vehicle body. Therefore, the vehicle velocity is limited to under 40 km/h to prevent image motion blur and achieve denser road scan. To enlarge the diversity of the dataset, raw data is collected on paved concrete and asphalt roads in urban and rural areas with various uneven conditions, covering about 30km of roads. Since there is the multipath effect of the RTK, only the segments with 1.4cm localization precision are preserved.

The single-frame LiDAR point cloud is still sparse, making fine-grained reconstruction challenging. Multi-frame fusion is conducted to accumulate nearby points. First, the points in nearby 4\textasciitilde6 frames are aligned to the same origin with the motion information, which is actually the motion compensation. Specifically, the translation and pose variation in the local ENU coordinate relative to the origin are interpolated for all points to be fused, after which the points are compensated and transformed into the original LiDAR coordinate. Then, the ICP \cite{121791} and improved forms further refine the fusion. The front and back frames are registered to the origin frame. We found that the geometric features near the road surface are lacking, and also, the road scenarios are so variable that the algorithms are not robust to all samples. To avoid extra noise and guarantee the dataset's quality, we manually fine-tune the ICP hyper-parameters by grid-search for every sample and pick the one with the highest alignment accuracy.

The average alignment errors in the road surface's horizontal and vertical directions are bounded by ±1.2cm. This error level guarantees the preservation of detailed road surface unevenness such as slight cracks and rocks.

\subsection{Dataset Contents}

The stereo images are rectified with calibration parameters. The fused dense road surface points are projected onto the image plane of the left camera (the one above LiDAR), and the points within the camera’s perspective are retained. Then, the depth and disparity values on the corresponding pixels are calculated.

It is costly to obtain massive fused point clouds of high accuracy. We pick out representative road conditions and establish the RSRD-dense containing 2800 pairs of stereo images, point cloud, depth and disparity maps. The rectified stereo images are saved in .jpg format with a save quality 100. The depth and disparity maps are saved in 16-bit .png format after multiplying 256. The point clouds are saved as .pcd files containing xyz values. Note that for removing the noise points that would corrupt the registration performance, the original point cloud may be cropped to retain only the nearby road surface area. 

All the separate image samples in the dataset can be utilized for recovering the single-frame road surface profile. Among them, there are 15 sequences each of 8 seconds long for localization and large-scale reconstruction. The raw pose, location, and velocity measurements are also attached and can be read with our dev-kit. Each sequence covers road of about 50 meters, which is applicable for multi-view stereo. 

For further enlarging the dataset scale and diversity, we provide another subset with sparse point cloud labels (i.e., RSRD-sparse). It contains about 13000 sample pairs annotated by motion-compensated single LiDAR frame, including 176 sequences with motion information. Since the points are quite sparse, this subset can be utilized for pre-training, weak-supervised or self-supervised learning. We do not recommend it for dense road surface perception applications.

\begin{table*}[t]
  \centering
  \setlength{\tabcolsep}{7pt}
  \caption{Comparison of the existing datasets with stereo images in AVs perception. 
  \emph{Road area} is the ratio of the road surface to the whole image. The \emph{KAIST Urban}, \emph{FordAV}, and \emph{Oxford Robot} datasets do not directly give the rectified stereo images.}
    \begin{tabular}{l|c|c|c|c|c|c|c|c}
    \toprule
     & \# samples & \tabincell{c}{Resolution} & \tabincell{c}{Stereo baseline \\ (cm)} & \tabincell{c}{Focal length \\ (px)} & \tabincell{c}{LiDAR acc. \\ (cm)} & \tabincell{c}{Road area \\ (\%)} & \tabincell{c}{GT ratio \\ (\%)} & \tabincell{c}{Disp. acc. \\  (px)} \\
    \midrule
    KITTI’12 \cite{Menze2015CVPR}  & \multicolumn{1}{c|}{389} & 1242$\times$375 & 54     & 707    & ±2 & 18.3   & 28.04  & 0.5 \\
    KITTI’15 \cite{Geiger2012CVPR}   & \multicolumn{1}{c|}{400} & 1242$\times$375 & 54     & 707    & ±2     & 20.6   & 19.72 & 0.6 \\
    Argoverse \cite{Chang_2019_CVPR}   & \multicolumn{1}{c|}{6624} & 2464$\times$2056 & 29.7   & 3757   & ±3     & 31.6   & 0.78   & 0.7 \\
    ApolloScape \cite{wang2019apolloscape}   & \multicolumn{1}{c|}{5165} & 3130$\times$960 & -- & -- & ~~±0.5   & 30.1   & 78.24 & 8.2 \\
    DrivingStereo \cite{yang2019drivingstereo}   & \multicolumn{1}{c|}{182188} & 881$\times$400 & 54     & 2061   & ±2     & 37.7   & 21.18  & 1.0 \\
    \midrule
    KAIST Urban \cite{8460834}    & --      & 1280$\times$560 & 47.5   & 775    & ±3     & 32.2   & -- & -- \\
    FordAV \cite{agarwal2020ford}   & --      & 1656$\times$860 & 52.9   & 945    & ±2     & 16.0     & -- & -- \\
    Oxford Robot \cite{RobotCarDatasetIJRR}  & --      & 1280$\times$960 & 24     & 983    & ±3     & 29.3   & -- & -- \\
    \midrule
    RSRD   & 2800 + 13672 & 1920$\times$1080 & 12     & 2022   & ±1     & 89.1   & 4.12   & 0.6 \\
    \bottomrule
    \end{tabular}%
  \label{comparison}%
\end{table*}%

\section{DATASET STATISTICS}

\subsection{Comparison}

To demonstrate the superiority of our dataset, we comprehensively compared the existing vision datasets for driving environment perception. Note that although point clouds and motion measurements are provided for large-scale reconstruction, we are more inclined to recover road surfaces with a single stereo pair for practical online applications. Recovering road profile by the binocular camera is more reliable than monocular depth estimation or multi-view stereo for practical applications. We do not consider the diverse stereo datasets based on robot platforms such as RELLIS-3D \cite{9561251} and S3E \cite{feng2022s3e}, since they have vastly different perspectives and scenarios from vehicles. We randomly extract 100 samples from every dataset, and the evaluation results on various metrics are listed in Table \ref{comparison}. 

The widely used KITTI dataset contains few samples in the stereo subset, based on which the performance of deep learning models to be developed cannot be ensured. The DrivingStereo has much more samples by collecting data in similar scenarios and road sections. The road ratio indicates the ratio of road area to the whole image. The existing datasets have low road ratios since they care the complete traffic environment. The label ratio is the percentage of pixels with ground-truth LiDAR points. More accurate and fine-grained estimations are achievable with denser label supervision. Nevertheless, this index is not directly comparable since it can be improved by reducing the image resolution in stereo rectification. Our RSRD still reaches 4.12\% even at 2M image resolution. The ApolloScape achieves extremely dense labels by fitting CAD models to cars and roads. Recovering the actual road profiles is almost impossible since the road surfaces are regarded as planes.

We also assessed the average disparity error, which is easier to implement than the LiDAR depth error. It is a comprehensive metric involving all the errors in sensor acquisition, fusion, and calibration processes. We pick corresponding pixels at different positions of the stereo images and calculate their errors between the LiDAR-measured disparity values. Although pixels are discrete, we pick continuous coordinate values with the maximum corresponding possibility. Our RSRD achieves an error of 0.6, which is generally equivalent to the KITTI. It outperforms all the other datasets as for the corresponding depth error because our cameras have a smaller value of focal multiplying baseline. The human-designed model fitting causes significant disparity errors in the ApolloScape. Also, the errors are inconsistent among samples, possibly because of the temporary loss of RTK. The Argoverse-stereo has higher errors at object boundaries, possibly because of poor motion compensation or joint calibration. 

The road condition diversity of the compared datasets is relatively poor as they focus on the whole traffic condition. The accuracy and label density do not satisfy the requirements of precise and dense road surface perception applications. By comprehensive comparison, our RSRD has superiority among all indexes and is a better alternative for road surface perception. 

\begin{figure}[t]
    \centering
    \includegraphics[width=1\linewidth]{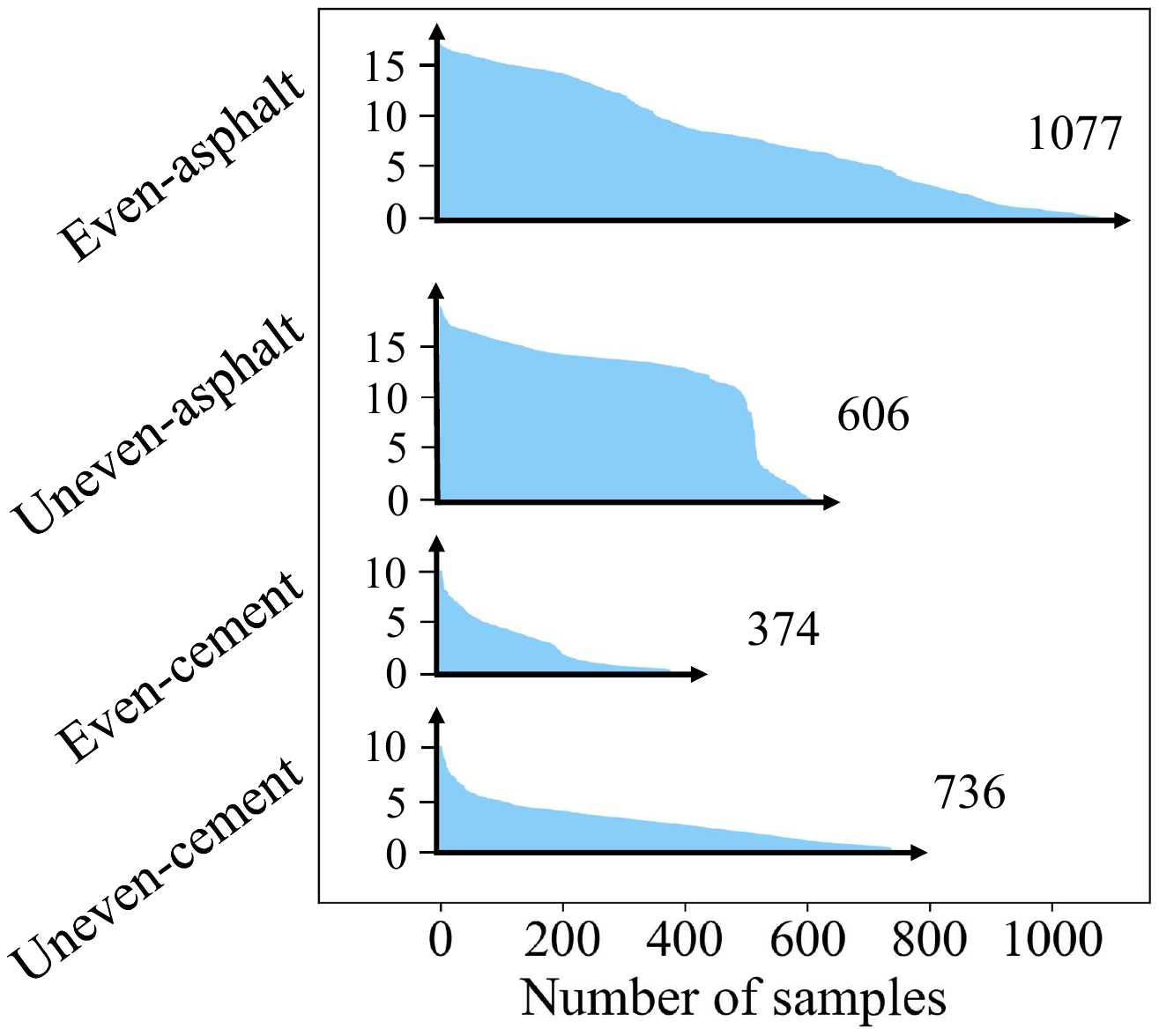}
    \vspace{-20pt}
    \caption{Counts of the road classes and corresponding texture indexes.}
    \label{texture}
\end{figure}

\subsection{Scenario Diversity}

As demonstrated above, we strive to enlarge the dataset coverage in experiments and data processing. For road surface perception, the dataset diversity can be indicated by road unevenness conditions and image texture richness. The counts of even and uneven conditions on asphalt and cement roads in the RSRD-dense are shown in Figure \ref {texture}. The uneven samples refer to roads with obvious cracks, bumps, or potholes that would cause vehicle jolt. For even road surfaces, the developed algorithms should recover approximately planer road profiles without much noise to prevent false positives. For uneven roads, they should accurately capture the geometric contour of bumps or potholes.

Unlike traffic objects, the road surface is texture-less without regular patterns. The coverage of roads with different texture features is crucial for the dataset. To assess the texture diversity, we calculate the dissimilarity index of the image gray-level co-occurrence matrix (GLCM), a commonly utilized metric to measure the local variation and uniformity of the image. It gives smaller values to more homogenous textures. The dissimilarity of the whole image is averaged among all pixels, and the sorted values in the descent order of every class are also shown in Figure \ref {texture}. 

The four subfigures illustrate that all the classes cover wide texture ranges, indicating roads of diverse aggregate properties, service ages, traffic flow, and crack patterns. The asphalt roads have richer textures than cement roads. Also, uneven road conditions have averagely larger dissimilarity values than even roads, as the texture changes sharply at the edge or inside of various road damages.

\begin{figure}[t]
    \centering
    \includegraphics[width=0.8\linewidth]{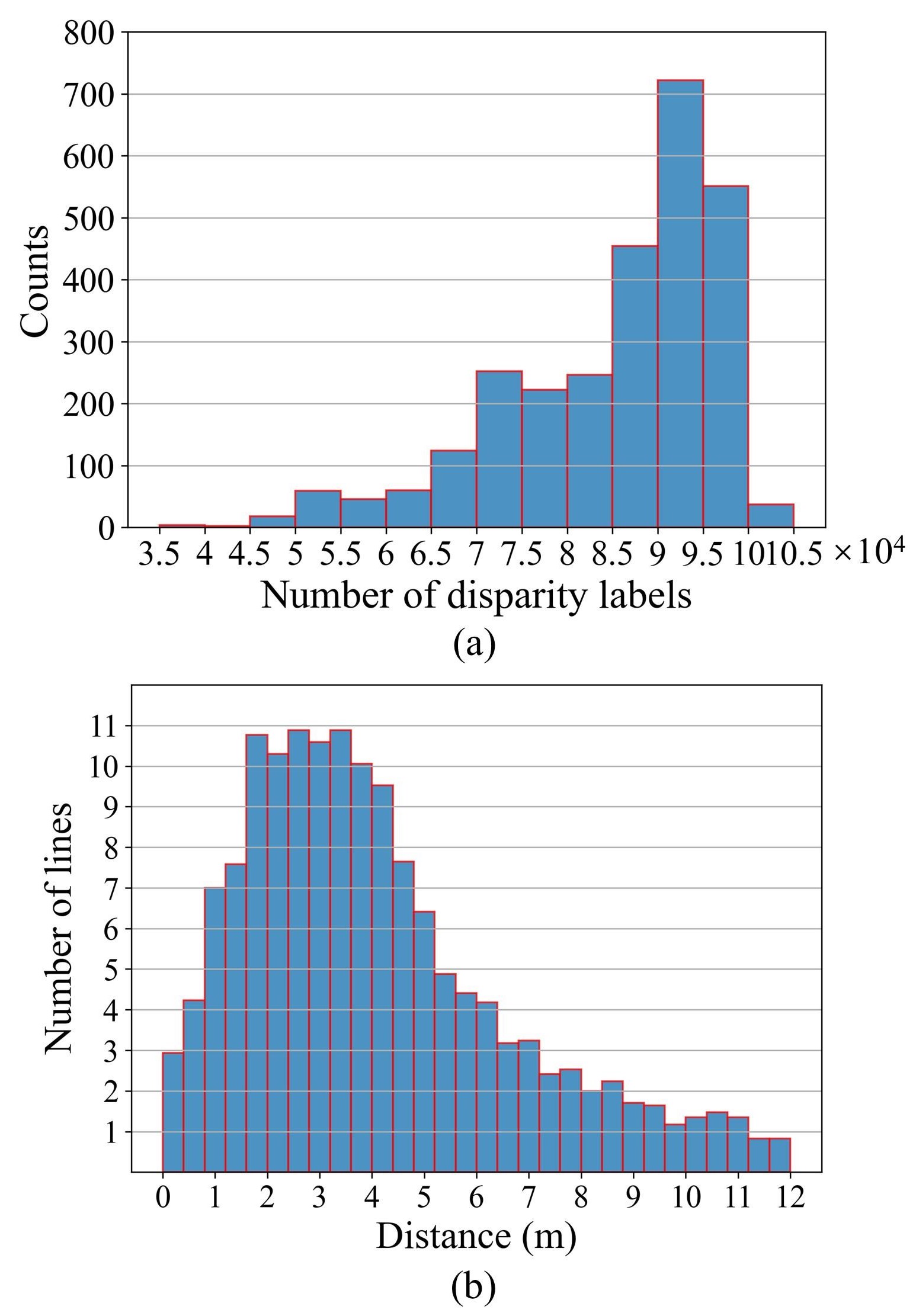}
    \vspace{-8pt}
    \caption{Density of point cloud labels. (a) Histogram of the number of disparity labels. (b) Number of LiDAR scanlines in the longitudinal direction.}
    \label{density}
    \vspace{-6pt}
\end{figure}

\subsection{Label Density}

Dense ground-truth labels are significant for accurate and fine-grained road profile recovery. We count the number of points in the left image of the RSRD-dense, and the histogram is shown in Figure \ref {density} (a). Most image samples have 70K\textasciitilde100K pixels with GT depth values, corresponding to the ratio between 3.4\%\textasciitilde4.8\%. Also, we evaluate the GT density along the longitudinal direction of the road surface. The number of LiDAR scanlines in every interval of 0.4 m is counted, as shown in Figure \ref {density} (b). Within six meters, the dataset ensures an average scanline every 10cm.  The reconstruction performance is expected to decrease from seven meters away since both the GT density and image definition are low.

\begin{table}[t]
  \centering
  \setlength{\tabcolsep}{1.5pt}
  \caption{Evaluation results with monocular depth estimation methods.}
    \begin{tabular}{p{6.9em}|c|c|c|c|c}
    \toprule
    Method & \multicolumn{1}{p{5em}|}{$\delta<1.25\uparrow$} & \multicolumn{1}{p{4em}|}{Abs Rel ↓} & \multicolumn{1}{p{3.5em}|}{RMSE ↓} & \multicolumn{1}{p{5.28em}|}{RMSE log ↓} & \multicolumn{1}{p{3.6em}}{Sq Rel ↓} \\
    \midrule
    AdaBins \cite{bhat2021adabins} & 0.998  & 0.016  & 0.150   & 0.023  & 0.005 \\
    NeWCRFs \cite{yuan2022newcrfs} & 0.993  & 0.033  & 0.294  & 0.044  & 0.017 \\
    BTS \cite{lee2019big}    & 0.998  & 0.019  & 0.172  & 0.026  & 0.006 \\
    SAN \cite{xu2018structured}   & 0.998  & 0.029  & 0.219  & 0.036  & 0.009 \\
    iDisc \cite{piccinelli2023idisc}  & 0.999  & 0.019  & 0.174  & 0.026  & 0.006 \\
    PixelFormer \cite{agarwal2023attention} & 0.998  & 0.019  & 0.176  & 0.026  & 0.006 \\
    LapDepth \cite{9316778} & 0.998  & 0.023  & 0.217  & 0.032  & 0.009 \\
    \midrule
    LapDe. (far) & 0.997  & 0.026  & 0.251  & 0.036  & 0.011 \\
    LapDe. (near) & 1.000      & 0.016  & 0.053  & 0.020   & 0.002 \\
    \bottomrule
    \end{tabular}%
  \label{mono}%
\end{table}%

\section{BENCHMARK}

Benchmark evaluations on typical computer vision tasks are conducted to verify the feasibility of RSRD in road reconstruction. As a baseline and prototype, we select state-of-the-art monocular depth estimation and stereo matching algorithms for evaluation. The RSRD-dense is split into train and test sets with 2500 and 300 samples, respectively. We do not split the RSRD-sparse. A demo video showing the visualization results is provided in the dataset website.

\begin{figure*}[t]
    \centering
    \includegraphics[width=1\linewidth]{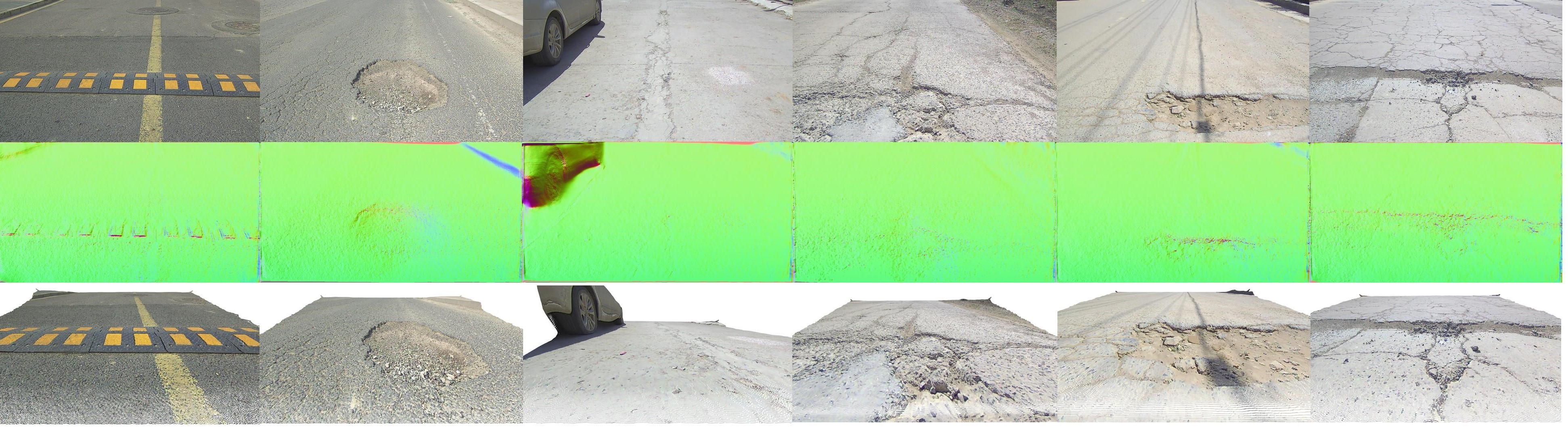}
    \caption{Inference results by monocular depth estimation method. From up to down: input RGB images, surface normal maps, and colored point clouds.
    For better visualization, we show the surface normal maps calculated from the depth maps. 
    }
    \label{mono_infer}
\end{figure*}

\subsection{Monocular Depth Estimation}
\noindent
\textbf{Training details.} We test seven depth estimation algorithms that ever achieved the SOTA performance. The full-resolution images of the left camera in RSRD-dense are taken as inputs. The maximum depth is set as 14 meters. The models are trained for ten epochs for fair comparison. No augmentations are implemented since they may introduce noise in interpolation. The batch size is set to fully utilize a single RTX 3090 GPU. All the other configurations are kept unchanged. 

\noindent
\textbf{Evaluation results.} We also adopt the commonly utilized metrics in-depth estimation to evaluate the models, as shown in Table~\ref{mono}. Benefiting from the high accuracy and dense point cloud labels, all the models achieve distinguished values on the metrics. However, the relative depth error around 2\% indicates an absolute error of 10cm at 5m depth, which will be bigger at farther distances. The accuracy is far from enough for practical applications since road surface vibrations are generally smaller than this level. The dataset is quite challenging and therefore, more advanced methods and models should be explored for more accurate estimation. Figure~\ref{mono_infer} shows examples of inference results. The road unevenness is not visually obvious in the depth maps since they have small amplitudes. Instead, we visualize the surface normal maps calculated from the depth maps. 

For more insights, we separately evaluated the metrics on the upper and lower half of the image. This evaluation method indicates the model performance at near and far distances. The results with the LapDepth model are shown at the bottom of Table \ref{mono}. The depth estimation accuracy at near distance enormously outperforms far distance, which is consistent with our analysis in the dataset description section. The road surface texture becomes coarse with the increase of preview distance. The models to be developed should consider the geometry and texture properties of the images.

\begin{table}[t]
  \centering
  \setlength{\tabcolsep}{1.6pt}
  \caption{Performance comparison of models trained on RSRD-dense and pretrained on RSRD-sparse.}
    \begin{tabular}{p{5.3em}|c|c|c|c|c}
    \toprule
    Dataset & \multicolumn{1}{p{4.945em}|}{$\delta<1.25\uparrow$} & \multicolumn{1}{p{4.055em}|}{Abs Rel $\downarrow$} & \multicolumn{1}{p{4.055em}|}{RMSE $\downarrow$} & \multicolumn{1}{p{5.28em}|}{RMSE log $\downarrow$} & \multicolumn{1}{p{4.055em}}{Sq Rel $\downarrow$} \\
    \midrule
    RSRD-dense & 0.998  & 0.023  & 0.217  & 0.032  & 0.009 \\
    Pretrained & 0.999  & 0.020   & 0.184  & 0.027  & 0.006 \\
    \bottomrule
    \end{tabular}%
  \label{pretrain}%
\end{table}%

\begin{table}[t]
  \centering
  \setlength{\tabcolsep}{9.5pt}
  \caption{Evaluation results with stereo matching methods}
    \begin{tabular}{p{7.5em}|c|c|c}
    \toprule
    Method & \multicolumn{1}{p{4.4em}|}{EPE (px)} & \multicolumn{1}{p{4.5em}|}{\textgreater1 px (\%)} & \multicolumn{1}{p{4.5em}}{\textgreater3 px (\%)} \\
    \midrule
    RAFTStereo \cite{lipson2021raft} & 0.450   & 8.139  & 1.157 \\
    ACVNet \cite{xu2022attention} & 0.354  & 4.885  & 0.100 \\
    IGEV \cite{xu2023iterative}   & 0.369  & 4.896  & 0.151 \\
    CFNet \cite{Shen_2021_CVPR}  & 0.333  & 3.276  & 0.063 \\
    GWCNet \cite{guo2019group} & 0.412  & 5.890   & 0.255 \\
    \bottomrule
    \end{tabular}%
  \label{stereo}%
\end{table}%

\subsection{Effects of Pre-training}
The RSRD-sparse subset has a much larger scale but with sparse point cloud labels, which is suitable for pretraining or weak-supervised learning. Taking depth estimation with LapDepth as an example, we validate the effectiveness of pretraining on the RSRD-sparse. The model is first trained on the sparse subset for five epochs and then transferred to the dense subset. The configurations and hyper-parameters are the same as those when training only on the dense subset. The comparison results in Table~\ref{pretrain} indicate that the pre-training promotes the model performance since the sparse subset's diversity is larger.

\subsection{Stereo Matching}
\noindent
\textbf{Training details.} We select five stereo matching methods to fit the dataset. The stereo pairs may be center-cropped since stereo matching for 2M resolution images burdens memory and computation consumption. The maximum disparity value is set as 128 for the cropped images. The five models are trained on the dense subset for five epochs.  

\noindent
\textbf{Evaluation results.} We evaluate the model performance with the end point error (EPE) calculated as the average absolute disparity error. The error ratios bigger than one and three pixels are also presented in Table \ref{stereo}. All the models perform well on the metrics. More than 95\% of pixels have an estimation error of less than 1 pixel. The disparity errors are around 0.4 pixels, which is at the sub-pixel level. Considering the camera's intrinsic and extrinsic parameters, the disparity error corresponds to a depth error of 4cm at 5m depth, which is smaller than that of monocular depth estimation. Recovering road profiles by stereo cameras is more promising than the monocular. 

\section{Discussions}
Our RSRD dataset is designed to enhance safety and comfort in autonomous driving, and we have demonstrated its effectiveness in tasks like depth estimation and stereo matching.
However impressive these advances and baselines for these tasks are in simple cases, the full promise of fine-grained road surface reconstruction has yet to be realized.
The task is very challenging due to the typically low-texture characteristic of roads.
Exploring the potential of methods such as NeRF~\cite{mildenhall2021nerf} and other Multi-View Stereo (MVS)~\cite{9863705}~\cite{9797764} techniques in this context hold promise as an exciting and under-explored topic for future research.
With our dataset and benchmarks, we aspire to facilitate and inspire advancements in this field, serving as a valuable resource for researchers and enthusiasts alike.
Our RSRD has no ethical or social issues on its own, except those inherited from autonomous driving.

\section{CONCLUSION}
We introduce RSRD, the first large-scale, high-accuracy, and high-resolution dataset for fine-grained road reconstruction. The RSRD comprehensively outperforms the existing datasets for road surface recovery applications. The high-accuracy labels guarantee that the models will not fit on noise. The 16K data pairs cover diverse road surface conditions, which have potential in real-vehicle applications. It is suitable for depth estimation, stereo matching, large-scale multi-view reconstruction, and also possibly, localization, and point cloud processing. Initial baseline evaluations verify that the dataset is challenging in recovering fine road structure and texture. Comprehensive road condition information can be obtained combined with the previously released RSCD.





\bibliography{IEEEabrv}

\end{document}